\documentclass{article}
\usepackage{ijcai23}

\usepackage{times}
\usepackage{soul}
\usepackage{url}
\usepackage[hidelinks]{hyperref}
\usepackage[utf8]{inputenc}
\usepackage[small]{caption}
\usepackage{graphicx}
\usepackage{amsmath}
\usepackage{amsthm}
\usepackage{booktabs}
\usepackage{algorithm}
\usepackage{algorithmic}
\usepackage[switch]{lineno}

\usepackage{xspace}
\usepackage{stfloats}
\usepackage{amssymb}
\usepackage{multirow}
\usepackage{todonotes}
\usepackage[inline]{enumitem}
\usepackage{float}

\newcommand{\model}{{\textsc{Indent}}\xspace}
\newcommand{\modelspeech}{{\textsc{Indent}}\xspace} 
\newcommand{\modeltext}{{\textsc{Indent-T}}\xspace} 
\newcommand{\zeroshot}{{\textsc{No-Train}}\xspace}
\newcommand{\gaussiannet}{{Gaussian Weighted Cross Attention}\xspace}

\title{Temporally Aligning Long Audio Interviews with Questions:\\ A Case Study in Multimodal Data Integration}

\author{
Piyush Singh Pasi$^1$
\and
Karthikeya Battepati$^1$\and
Preethi Jyothi$^{1}$\and
Ganesh Ramakrishnan$^1$\and \\ 
Tanmay Mahapatra$^2$\and
Manoj Singh$^2$
\affiliations
$^1$Indian Institute of Technology, Bombay\\
$^2$CARE India Solutions for Sustainable Development\\
\emails
\{piyushsinghpasi, karthikeyabatte, pjyothi, ganesh\}@cse.iitb.ac.in,
\{tanmay, manojks\}@careindia.org
}

\begin{document}

\maketitle

\begin{abstract}
The problem of audio-to-text alignment has seen significant amount of research using complete supervision during training. However, this is typically not in the context of long audio recordings wherein the text being queried does not appear verbatim within the audio file. This work is a collaboration with a non-governmental organization called CARE India that collects long audio health surveys from young mothers residing in rural parts of Bihar, India. Given a question drawn from a questionnaire that is used to guide these surveys, we aim to locate where the question is asked within a long audio recording. This is of great value to African and Asian organizations that would otherwise have to painstakingly go through long and noisy audio recordings to locate questions (and answers) of interest. Our proposed framework, \model, uses a cross-attention-based model and prior information on the temporal ordering of sentences to learn speech embeddings that capture the semantics of the underlying spoken text. These learnt embeddings are used to retrieve the corresponding audio segment based on text queries at inference time. We empirically demonstrate the significant effectiveness (improvement in R-avg of about 3\%) of our model over those obtained using text-based heuristics.  We also show how noisy ASR, generated using state-of-the-art ASR models for Indian languages, yields better results when used in place of speech. \model, trained only on Hindi data is able to cater to all languages supported by the (semantically) shared text space. We illustrate this empirically on 11 Indic languages.
\end{abstract}

\section{Introduction}

Audio surveys and oral interviews are routinely used as a means for data collection in many parts of the world~\cite{audiosurvey}, \cite{reichmann2010does}. Apart from ease of use, audio surveys are also very inclusive since they naturally allow for data to be collected from illiterate or physically challenged individuals~\cite{heinritz2022surveying}. Several audio surveys by governmental and non-governmental organizations are conducted with the aim of improving social outcomes such as health and education. These surveys are typically accompanied by predefined questionnaires that guide the interviewer. Temporally aligning questions to where each question was asked in an interview becomes very important for organizations collecting the data, so as to quickly retrieve answers to relevant questions from long audio files.

\begin{figure}[t]
    \centering
    \includegraphics[width=\columnwidth]{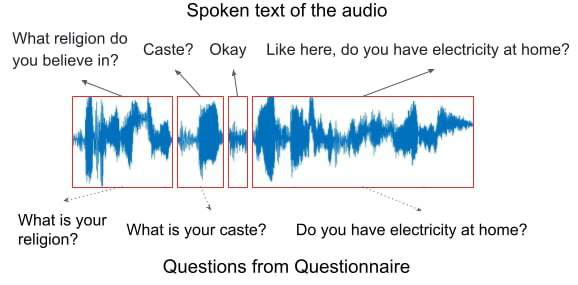}
    \caption{Illustration of speech in audio and relevant questions}
    \label{fig:illustration}
\end{figure}

In this work, we use long audio health surveys collected from mothers with young children residing in rural areas of Bihar, India. This data was collected by a non-governmental organization CARE India~\cite{CARE} as part of their larger effort towards improving maternal and neonatal health. The interviews are typically long, spanning more than 40 minutes on average, and are based on a fixed questionnaire designed by CARE. Given a written question drawn from the questionnaire, our main goal is to extract a short audio snippet from the long recording that matches the given question. This task is of utmost importance to organizations such as CARE who would otherwise have to resort to tedious manual searches through these long audio recordings to retrieve answers to various questions.

Most prior work on cross-modal segment retrieval relies on strong supervision in the form of temporally-aligned text annotations \cite{anne2017localizing}, \cite{gao2017tall}, \cite{liu2018cross}, \cite{xuarticle}, \cite{gupta2021clsril}, \cite{javed2022towards}. For long audio files, creating such time-aligned annotations using human annotators would require significant amount of time and money. Since we are working with audio surveys derived from a fixed sequence of questions in a questionnaire, we use a weaker form of supervision involving larger audio segments paired with sets of temporally ordered questions. We present a new architecture \model\footnote{\model can be expanded as ``al{\bf I}gning lo{\bf N}g au{\bf D}io int{\bf E}rviews a{\bf N}d  ques{\bf T}ions".}, that is trained using such weak supervision of larger audio segments and sets of questions. \model uses an attention-based mechanism and is trained with contrastive losses across audio-text modalities to ground each question in the larger segment with its corresponding audio.

Our task of audio segment retrieval using textual queries is particularly challenging for the following reasons:
\begin{enumerate*}
\item The audio surveys with the young mothers are conducted in noisy home environments, thus resulting in very noisy audio recordings.
\item Questions from the questionnaire are not asked verbatim and are typically paraphrased by the interviewer when posed to the mothers. Figure~\ref{fig:illustration} illustrates this phenomenon. 
\item While the questions appear in formal text in the official language of Bihar, {\em i.e.}, Hindi, the audio recordings could contain variations that are typical of the local dialects spoken by the interviewees.
\end{enumerate*}

To summarize, our main contributions are: (i) We tackle a high-utility problem of locating short audio segments from within long audio surveys based on textual queries. These audio surveys are collected by CARE India in real environments and present many interesting challenges that we outline above. The dataset is enriched with fine-grained annotations on the evaluation splits and coarse-grained annotations on the training data that is sufficient to enable weak supervision. (ii) We design \model, a new model for text-to-audio retrieval from within long audio surveys given a question in text. \model is trained using weak supervision consisting of longer audio segments and sets of temporally ordered questions. 

The problem of question-to-audio retrieval can also serve as an intermediate step for various downstream tasks such as audio-driven form filling and audio-driven question answering, by extracting answers from the survey recordings once the relevant question is isolated. We leave such extensions to future work and focus on the challenging problem of retrieving audio segments from long audios based on questions in text.

\section{Related Work}

Most relevant to our work are (i) approaches that focus on temporal sentence grounding in videos, where the task is to retrieve a video that semantically corresponds to a query in natural language, and (ii) audio text alignment for Automatic Speech Recognition (ASR).

\subsection{Video Moment Retrieval Using Text Queries} 

Prior work has explored various approaches for the task of temporally grounding captions or descriptions in videos in both fully and weakly supervised settings~\cite{survey}. Some of the prominent approaches are discussed below. 
Moment Context Networks (MCN)~\cite{anne2017localizing} learn a shared embedding for video temporal context features ({\em i.e.}, video features integrated with temporal endpoint features that indicate when a moment occurs in a video) and LSTM-based language features. The CTRL architecture \cite{gao2017tall} constitutes a temporal regression network to produce alignment scores and location offsets using combined representations for visual and text domains. A language-temporal attention network has also been proposed~\cite{liu2018cross} to learn word attention based on the temporal context information in the video. Temporal boundary annotations for sentences have been used to train a segment proposal network using attention~\cite{xuarticle}. All these approaches assume the fully supervised setting. A Text-Guided Attention (TGA) model was proposed for video moment retrieval using text queries in a weakly supervised setting~\cite{mithun2019weakly}. However, TGA does not utilize the temporal order of sentences unlike our approach, \model 
that employs a Gaussian-weighted attention to leverage this temporal information. 

\subsection{Audio-Text Alignment For Speech Recognition}

Most of the widely used ASR models that generate the best transcripts for various languages require large amounts of labeled transcripts for training~\cite{chadha2022vakyansh,javed2022towards}. Methods proposed for low resource settings \cite{anguera2014audio} also expect access to some annotated transcripts during training, which are not available in our setting. Models that use CTC or sequence-to-sequence loss functions \cite{synnaeve2019end,lopez2022iterative} require audio files with their corresponding transcripts to be used during training. Such models cannot be employed with our CARE India dataset, since we do not have any transcribed speech and the questions asked in the audio recordings are only semantically similar to (and often, paraphrases of) the questions listed in the questionnaire.


Recently, a few methods have been proposed for speech recognition using unaligned speech and text ~\cite{ao2021speecht5,chen2022maestro} and generating speech embeddings that capture the semantics of the underlying spoken text by aligning both text and speech embedding spaces using unsupervised techniques~\cite{chung2018unsupervised,chen2018towards}. These models require word-level audio segments for semantically rich speech embeddings, which are generally not easily available or extractable. Other prior work require very large amounts of unpaired speech and text and do not currently serve many Indian languages~\cite{ao2021speecht5,chen2022maestro}. 

\section{CARE India Dataset}
\label{sec:data}

We create a new ``CARE India Dataset" that contains speech interviews conducted by the CARE India Organization\footnote{\url{https://bihar.care.org/}}. These interviews focus on analyzing maternal health and empowering women in rural areas in the state of Bihar, India. Each interview is approximately 40 minutes long and is recorded in the household premises of the interviewee, thus yielding very noisy recordings. The interviewer refers to a questionnaire as a guide for the duration of the interview. Searching for a specific question from the questionnaire within the noisy long audio recording is not the only challenge. Questions are not asked verbatim during the course of the conversation and could be paraphrased. The questions are also drawn from a diverse pool including multiple-choice type, numerical and subjective questions. All these factors add to the overall difficulty of isolating the correct audio segment that maps to a given question. CARE India (and other similar organizations) could greatly benefit from an automated solution that given a question in text can identify roughly where it appeared in a long audio recording. Without such a tool, volunteers would have to tediously go through long and noisy audio recordings to search for where a question appeared in it. This work is our first step towards building such a tool. We present a new model \model that can be trained on longer audio segments spanning multiple questions and identify potential audio segments within a test audio recording that could be matched with a test query. 

Each interview in the train set is annotated at the segment-level. A segment contains $m$ consecutive questions and is of variable duration. Segments do not overlap or share questions. If a question begins within a segment,  it would also end in that segment. We choose to set $m = 5$, since this granularity gives us segments that are not too long (which would make learning difficult) and not too short (which would otherwise overwhelm the human annotators). For each audio segment containing five questions, we annotate the start of the first question  and the end of the fifth question. Our development and test sets have fine-grained annotations with start and end timestamps for each question. However, such a per-question annotation takes much longer. We find that annotating train files takes around 4 hours per audio recording, while it takes more than 6 hours to annotate test files from a similarly-sized audio recording. Detailed dataset statistics for our train, development and test sets are presented in Table~\ref{tab:data-stats-hindi}.

\begin{table}[htpb]
\centering
\resizebox{0.8\columnwidth}{!}{%
\small
\begin{tabular}{|l|c|c|}
\hline
\textsc{Train set}                                  &                                                         \\ 
Total no. of interviews                    & 34                                   \\
Total no. of segments                       & 1223                                \\
Avg. no. of segments in an interview                  & 35.97                       \\
Avg. segment duration (sec)                   & 54.64                            \\
Avg. no. of chunks per segment                   & 14.31                        \\
Avg. chunk duration (sec)                      & 1.7                         \\

Avg. interview duration\textsuperscript{\textdagger} (min) & 42.1  \\
Total Train set duration\textsuperscript{\textdagger}  (hour) & 23.85                                                \\
Total no. of questions asked                           &            6115                                              \\
Avg. no. of questions asked in an interview                          & 179.85                                                      \\ \hline
\textsc{Dev set}                                    &                                                       \\ 
Total no. of interviews                    & 3                                  \\
No. of chunks in an interview                  & 634.33                      \\
Avg. interview duration\textsuperscript{\textdagger} (min) & 34.2  \\
Total Dev set duration\textsuperscript{\textdagger} (hour) &           1.71  \\
Avg. chunk duration  (sec)                  & 2.04                         \\
Total no. of questions asked                           &               538                                          \\
Avg. no. of questions asked in an interview                          & 179.33                                                        \\
Avg. question duration (sec)                        &              2.87                                           \\ \hline
\textsc{Test set}                                    &                                                         \\ 
Total no. of interviews                    & 5                                   \\
No. of chunks in an interview                  & 603.4                           \\
Avg. interview duration\textsuperscript{\textdagger} (min) & 36.28  \\

Total Test set duration\textsuperscript{\textdagger}   (hour)          & 3.02                                           \\
Avg. chunk duration (sec)                    & 1.96                     \\
Total no. of questions asked                           &             629                                            \\
Avg. no. of questions in an interview                          &      125.8                                                   \\
Avg. question duration (sec)                        &                   1.78                                      \\ \hline
No. of question in fixed questionnaire $Q$ & 1555  \\ \hline
\end{tabular}%
}
\caption{CARE India Dataset Statistics, {\textdagger} denotes duration without any preprocessing}
\label{tab:data-stats-hindi}
\end{table}

\section{\model: Architecture and Methodology}
We formally define our problem as follows. Given a question $q$ drawn from a fixed questionnaire $Q$\ and an audio interview $f$ within which  $q$ is asked, we want to retrieve the audio segment $s \in f$ that exactly matches $q$. On an average, $f$ is more than 40 minutes long. In Table~\ref{tab:data-stats-hindi}, we provide more detailed statistics on the data distribution. As described in Section~\ref{sec:data}, during training, we assume access to longer audio segments containing $m=5$ questions. These segmentations are manually annotated. At test time, we extract non-overlapping audio segments of fixed duration from the test audio interview $f$ and identify the segment that is the best match for a given question using our trained \model. 


\paragraph{Overview of Workflow.}  \model consists of three main components:
\begin{enumerate*}
    \item Speech Encoder
    \item Question Encoder
    \item Gaussian-weighted Cross-attention Network.
\end{enumerate*}
The speech encoder consists of audio preprocessing and feature extractor modules. We use an audio preprocessing module to remove noise from the speech interviews and break segments into smaller chunks of  roughly 2 seconds duration each. We extract speech features at the chunk-level. The question encoder extracts question features at the sentence-level. Hence, the time resolution of the audio chunk features and the question features are very different ({\em c.f.} Figure~\ref{fig:illustration}). Using both audio chunk and question feature sequences as inputs, we try to align chunks with questions and simultaneously bridge the modality gap between speech and text in a shared space using a \gaussiannet module and contrastive learning. At inference, for a speech interview, we rank segments using the dot-product scores of the chunks and the given question. In Figure~\ref{fig:e2e-arch}, we present a schematic diagram of the overall architecture of \model. Next, we  elaborate on each of the model components.

\subsection{Speech Encoder}
The audio recordings contain speech both from the interviewer and the interviewee; the latter is typically very faint and more difficult to isolate. As mentioned in Section~\ref{sec:data}, the audio recordings are annotated with start and end times of large segments containing $m=5$ questions each. Within each of these segments, we are only interested in parts corresponding to the interviewer. To isolate these parts, we attempted several techniques including voice activity detection (VAD)~\cite{vad}, speaker diarization~\cite{bredin2020pyannote} and source separation techniques~\cite{speechbrain}. After qualitatively analyzing the respective outputs, we found VAD to yield the best quality outputs. Using VAD, we split each annotated segment $s$ into $n_s$ chunks $\{c^s_i\}_{i=1}^{n_s}$. This results in a varying number of chunks across segments. We note here that the number of chunks $n_s$ can be higher than the number of questions $m$ in a segment $s$.  The microphone often fails to adequately capture the interviewee's voice, and we therefore assume (after qualitative validation) that the chunks after VAD would primarily contain the interviewer's speech.

Once we have the chunk sequence $\{{c^s_i}\}_{i=1}^{n_s}$, we use a frozen pretrained speech encoder (wav2vec2.0 \cite{baevski2020wav2vec}) to extract features for chunks $\{\boldsymbol{c}^s_i\}_{i=1}^{n_s}$ where ${\boldsymbol{c}^s_i} \in {R}^{T_i \times d'}$, $T_i$ is the number of time frames for chunk $c_i^s$. The speech encoder produces a $d'$-dimensional feature vector for each speech frame in $c_i^s$. To reduce ${\boldsymbol{c}^s_i}$ to one aggregate feature vector per chunk, we apply a 1D convolution and take the mean across all $T_i$ features. Next, we project these chunk features to a $d$-dimensional shared space using a linear layer. We then apply self-attention over the segment so that each chunk can gather context from other chunks. We also add skip connections wherever appropriate. The left block in Figure~\ref{fig:e2e-arch} represents the speech encoder with layers and skip connections. More formally:

\begin{align*}
    \{{{\boldsymbol{\hat{c}}}}_i^s\}_{i=1}^{n_s} &= \mathrm{MeanPool}(\mathrm{Conv1D}\left(\{\boldsymbol{c}^s_i\}_{i=1}^{n_s}\right))
    \\
    C^s &= {\{{{C_i^s}}}\}_{i=1}^{n_s} = 
  \mathrm{Linear} (\{{{\boldsymbol{\hat{c}}}}_i^s\}_{i=1}^{n_s})
    \\
    \boldsymbol{C}^s &= \{{\boldsymbol{C}_i^s}\}_{i=1}^{n_s} =  \mathrm{SelfAttn}({C^s}, {C^s}, {C^s})
\end{align*}
where ${C_i^s} \in \mathbb{R}^d$ is the chunk feature after a $\mathrm{Linear}$ layer, $\mathrm{SelfAttn}(k,q,v) = \mathrm{softmax}(\frac{k.q^{\intercal}}{\sqrt{d}})v$, $q, k, v \in \mathbb{R}^{n_s \times d}$ and $\boldsymbol{C}_i^s$ is the chunk feature after self-attention.

\subsection{Question Encoder}
\begin{figure}
    \centering
    \includegraphics[width=1\linewidth]{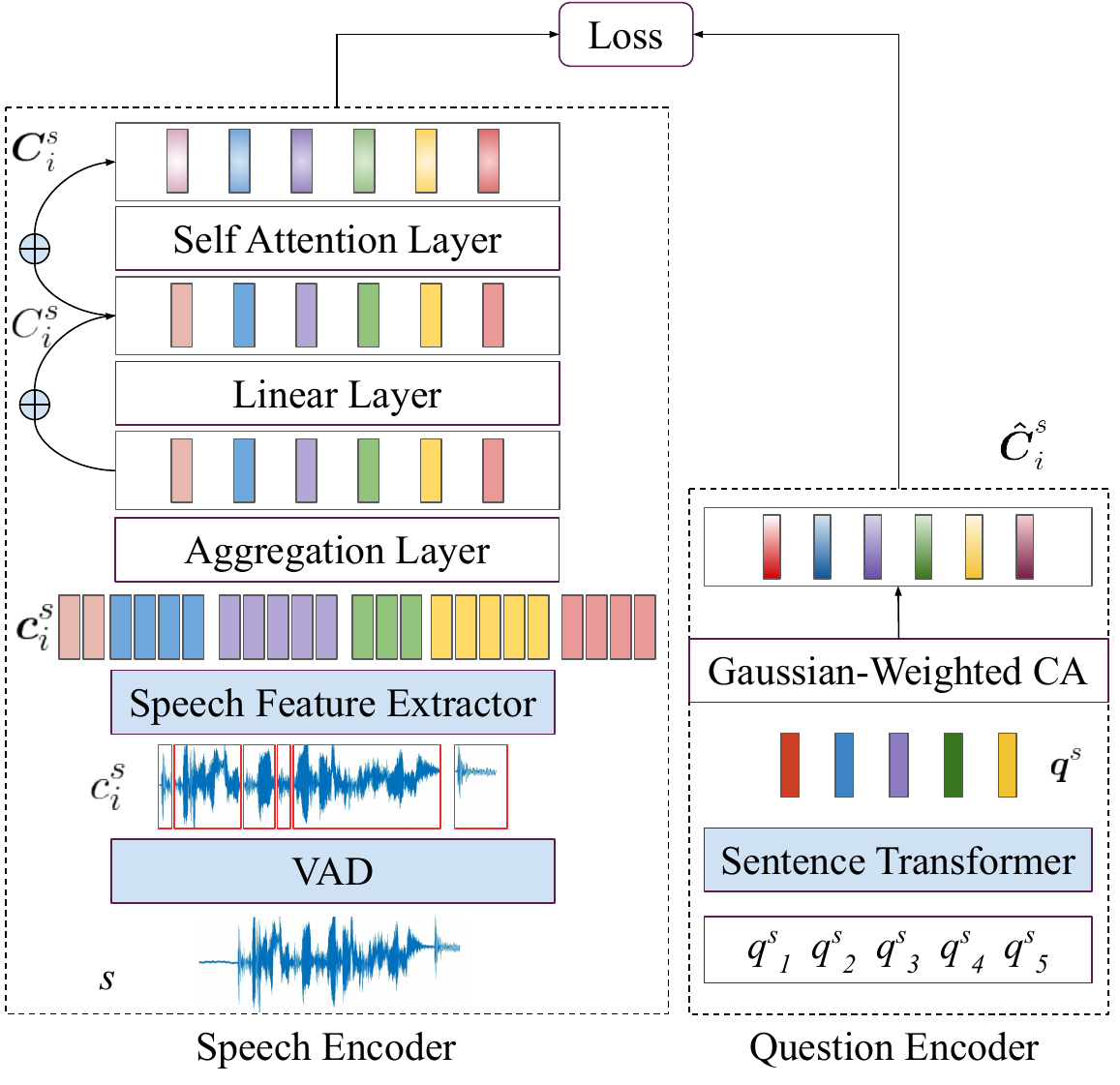}
    \caption{End-to-End \model Architecture. Modules filled with blue color are frozen.}
    \label{fig:e2e-arch}
\end{figure}

We use a frozen pretrained sentence transformer network to extract a $d$-dimensional feature vector. During training, for each segment $s$ we have $m$ questions, and hence we compute ${\boldsymbol{q}^s = [\boldsymbol{q}^s_1, \cdots, \boldsymbol{q}^s_m]}$ where $\boldsymbol{q}^{s}_{j} \in {\mathbb{R}}^{d}$. We opt for sentence-level features for textual questions since the spoken and textual questions are semantically similar at the sentence-level but need not have any correspondence at the word-level ({\em e.g.}, when the spoken question is a paraphrase of the written question). Thus, extracting word-level features from the textual questions can lead to sub-optimal alignments. We also note here that we are doing weakly supervised training with no access to temporal boundaries but only the order of the occurrence of questions within a segment. This motivates our choice of using sentence-level features. The right block in Figure~\ref{fig:e2e-arch} shows the question encoder. It is important to keep the question encoder frozen since the pretrained sentence transformer is trained with massive amounts of text data in comparison to our training data. We only want to learn  effective semantic alignments from the spoken utterances to the already pretrained sentence embeddings.

For each segment $s$, we now have speech chunk features $\boldsymbol{C}^s$ and textual features from the questions $\boldsymbol{q}^s$ and we need to learn a cross-modal alignment between these two feature sequences. Towards this, we propose a Gaussian-weighted Cross Attention scheme. 

\subsection{Gaussian-weighted Cross Attention}
\label{sec:contrastive}

Given $\boldsymbol{C}^s$ and $\boldsymbol{q}^s$, we aim to learn an alignment between $\boldsymbol{q}^s$, representing $m$ sentence embeddings and  $\boldsymbol{C}^s$, representing $n_s$ chunk embeddings. We propose a Gaussian-weighted cross attention module with a contrastive learning objective \cite{chen2020simple} in order to learn this alignment. Typically, $n_s > m$ implying that a question spans more than one speech chunk. To align a chunk $\boldsymbol{C}_i^s$ with a question $\boldsymbol{q}_i^s$, we consider an ``anchor chunk" feature $\hat{\boldsymbol{C}_i^s}$ using $\boldsymbol{q}^s$. While chunk is anchored to one question, the neighboring questions can also provide context. We assume some consecutive chunks $\boldsymbol{C}^s_{a:b}$ combine to form the question $\boldsymbol{q}_i^s$, but we do not know the temporal boundaries of $\boldsymbol{q}_i^s$ (given our weakly-supervised setting). Further, speech features $\boldsymbol{C}_i^s$ are not semantically rich to guide us with any weak boundaries. However, we know that each chunk unambiguously belongs to a single underlying question. Hence, we represent each chunk $\boldsymbol{C}_i^s$ as a linear combination of $\boldsymbol{q}^s$ but anchor the chunk $\boldsymbol{C}_i^s$ to a single $\boldsymbol{q}_i^s$ by making one of the weights high and the rest much smaller to accumulate some neighboring context. A Gaussian distribution with a moving mean and standard deviation serves our purpose. 

Start-of-segment chunks and end-of-segment chunks map to questions $\boldsymbol{q}_1^s$ and $\boldsymbol{q}_m^s$, respectively, with high probability. 
Hence, pivoting on $n_s$, we move the Gaussian mean and vary the standard deviation as a function of the position of the chunk within the segment:
\begin{align}
    \mu_i &= \frac{(i-1)(m-1)}{n_s -1}\\
    \sigma_i &= \alpha \cdot \mathrm{min}(i-1, n_s-i)\label{eq:sigma} 
    \\
    \mathcal{G}_i &= \mathrm{Gaussian}(\mu_i, \sigma_i)
    \label{eq:moving-G}
\end{align}
\noindent where $\mu_i$ and $\sigma_i$ are the mean and standard deviation used to generate the Gaussian distribution $\mathcal{G}_i$ for chunk $c_i^s$ and $\alpha$ is a scaling factor. 
From Eqn~\ref{eq:moving-G}, we see that for each position $i$, the mean shifts by $\frac{m-1}{n_s - 1}$ from the previously calculated mean. 

\begin{figure}[h]
    \centering
    \includegraphics[width=0.8\columnwidth]{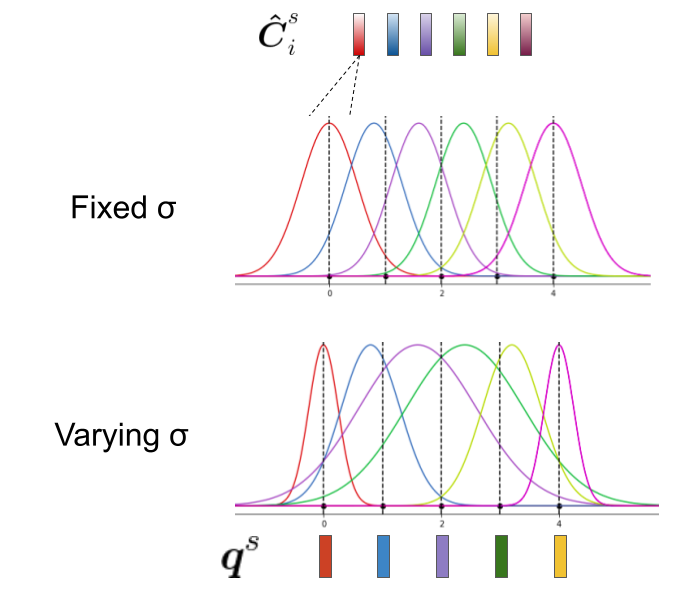}
    \caption{Gaussian-weighted Cross Attention (CA). Gaussian distribution of weights ($a^i_j$) for fixed and varying $\sigma$ cases. We sample at weights at each black-dot ($0, \cdots, m-1$) and the weight value is intersection of the Gaussian distribution and the dotted line which results in very peaky distribution for peripheral chunks \& vice versa.}
    \label{fig:gaussian}
\end{figure}

We sample values $\{a^i_j\}_{j=1}^{m}$ from $\mathcal{G}_i$ as shown in Fig.~\ref{fig:gaussian}, apply min-max normalization and generate an anchor chunk ${\hat{\boldsymbol{C}_i^s}}$ using $\boldsymbol{q}^s$ for each chunk $\boldsymbol{C}_i^s$.  Hence, $\{a^i_j\}_{j=1}^{m}$ has a peak value $a^i_j$ and also has some weight allocated to the other questions. 
Thus, we generate a representation corresponding to each audio chunk $\boldsymbol{C}_i^s$ in the shared semantic space as $\boldsymbol{\hat{C}}_i^s = \sum_{j=1}^m a^i_j {{q_j}^s}$. Since $\boldsymbol{\hat{C}}_i^s$ and $\boldsymbol{C}_i^s$ are both chunk-level representations, we apply a contrastive objective between them to bridge the modality gap. Since the question encoder is frozen, contrastive learning facilitates the speech encoder to better align with the text-based sentence embeddings.

\paragraph{Dynamic In-Audio Negative Sampling.}\label{neg-sampling} For contrastive learning, we need negative chunks. Specifically, negative chunks  from the same interview are desirable to prevent the model from learning easy solutions by merely exploiting speaker characteristics. During training, we create groups by randomly shuffling all segments in the speech interview and making groups of $n$ segments. If the number of segments in the audio interview is not divisible by $n$, we also create a group with the last $n$ segments in order to ensure that all the segments are covered during training. For each chunk in segment $s_x$, the corresponding negative chunks will be from segment $s_y$ ($x \neq y$) in the same group $g$. Chunks from all such $s_y$ are potential negatives for each chunk of $s_x$. We denote the number of all potential negatives for $s_x$ by $k_x$. To keep the number of negatives for each chunk constant inside a batch, we first compute the smallest $k_x$ across $s_x$'s in a batch and randomly sample those many chunks as negative examples for each chunk in the batch. We also explore data augmentation by performing reshuffling of all segments to create more new groups. The number of times we perform data augmentation is denoted by the hyperparameter $D$. Reshuffling is also performed across epochs.\\

\paragraph{Training Objective.} We define the following contrastive loss~\cite{chen2020simple} objective:
\begin{align*}
    L_{\text{st}} &= \frac{-1}{|C|} \sum_{C} \mathrm{log} \left( \frac{
    \mathrm{exp}(
    {\boldsymbol{C}_i^s}^{\intercal} \hat{\boldsymbol{C}_i}^s
    )
    }{\sum_{j \in n_g} \mathrm{exp}(
    {\boldsymbol{C}_j^s}^{\intercal} \hat{\boldsymbol{C}_i}^{s'}
    )} \right)
    \\
    L_{\text{ss}} &=  \frac{-1}{|C|} \sum_{C} \mathrm{log} \left( \frac{
    \mathrm{exp}(
    {\boldsymbol{C}_i^s}^{\intercal} {\boldsymbol{C}_i}^{s}
    )
    }{\sum_{j \in n_g} \mathrm{exp}(
    {\boldsymbol{C}_j^s}^{\intercal} {\boldsymbol{C}_i}^{s'}
    )}  \right)
\end{align*}
where $n_g$ is a union of $i$ and dynamically sampled negative chunk indices (as described earlier), $s, s' \in g, s=s'$ if $j=i$, $C$ denotes all chunks in a batch; $L_{\text{st}}$ aligns the speech and text spaces while $L_{\text{ss}}$ regularizes the network by explicitly encouraging negative speech chunks $\boldsymbol{C}_j^s$ ($j \in n_g, j \neq i$) to be far apart from $\boldsymbol{C}_i^s$ in the shared space.
\model is optimized over the loss $L = L_{\text{st}} + L_{\text{ss}}$. We apply the label smoothing technique~\cite{reed2014training} that forces \model to lower the confidence of the positive chunk by redistributing a small fraction of its probability mass uniformly to all the negative chunks. (In the experiments, we set positive probability to be 0.95 and redistribute 0.05 probability mass evenly among all the negatives.)
\section{Experiments and Results}\label{sec:exp}

\subsection{Experimental Setup}

\paragraph{Question Encoder.} We use pretrained Language-agnostic BERT Sentence Embedding (LaBSE)~\cite{labse} as sentence embeddings. For each question $q$, LaBSE produces a $768$-dimensional embedding. 
Since we keep the question encoder frozen, a helpful consequence of using LaBSE is that we can use questions in any language that are well-aligned in the LaBSE embedding space. In Table~\ref{tab:ab-lang}, we present experiments on questions in multiple Indian languages.

\paragraph{Speech Encoder.} We use a pretrained voice activity detector (VAD)~\cite{vad} to split segment $s$ into $n_s$ chunks, $\{c_1^s,\cdots,c_{n_s}^s\}$. To extract the chunkwise speech features $\boldsymbol{c_i}^s$, we use a state-of-the-art multilingual model IndicWav2vec-Hindi~\cite{javed2022towards}. 
Each chunkwise speech vector is a $1024$-dimensional feature vector. 
To aggregate, we apply a convolution layer with 1 filter per channel ({\em i.e.}, a total of 1024 filters) of receptive-field 20 and stride 2, followed by GELU activation \cite{hendrycks2016gaussian} and dropout and a final mean pooling on $\boldsymbol{c}_i^s$ to derive $\boldsymbol{\hat{c}}_i^s$. We then linearly project $\boldsymbol{\hat{c}}_i^s$  to a $768$-dimensional vector to generate $C_i^s \in C^s$. We subsequently apply self attention on $C^s$ to generate $\boldsymbol{C}^s$. We present experiments with different speech feature extractors in Section~\ref{subsec:result}. 

\paragraph{Training.} We train \model end-to-end keeping VAD~\cite{vad}, Question Encoder LaBSE~\cite{labse} and speech feature extractor IndicWav2Vec-Hindi~\cite{javed2022towards} modules frozen using the loss defined in Section~\ref{sec:contrastive}. We use the Adam optimizer with a learning rate 3e-4, a step scheduler with step-size 10 and decay factor gamma 0.1.
After hyperparameter tuning (described in Section~\ref{subsec:result}), we set batch size $|B|=4, n=4, D=2, \sigma=0.5$ and train for 40 epochs. 

\paragraph{Model Variants.} We refer to the complete model with the speech encoder and question encoder, trained using contrastive losses, as \model. To compare how  \model performs with text instead of speech, we train \modeltext by replacing the speech feature extractor in the speech encoder with ASR predictions for which we extracted LaBSE embeddings. \zeroshot is a learning-free technique that computes a simple matching based on dot-products between LaBSE embeddings of questions with LaBSE embeddings of ASR predictions from the speech.

\paragraph{Evaluation \& Metrics.} During inference, we create fixed-chunk-size segments since we do not have segment boundaries during inference. We need segment information for local context; removing self-attention within a segment deteriorates the performance significantly. We set  segment-size to 14 (average number of chunks in the train set). We compute dot-product score $q^\intercal \boldsymbol{C}_i^s$ for the given question $q$ across all chunks $\boldsymbol{C}_i^s$. We rank each segment $s$ using $\max_{i=1\cdots n_s} (q^\intercal \boldsymbol{C}_i^s)$ (higher score is better). We evaluate using the recall@k (R@k) metric. If the ground truth segment $s$ of the  question $q$ appears in the top $k$ highest scoring segments,  the algorithm is considered to have retrieved the correct segment for R@k. Our reported R@k numbers are a mean across interviews of the ratio of the number of questions correctly retrieved to the total number of questions asked $\left( \frac{\mathrm{no.\ of\ correctly\ retrieved\ questions}}{\mathrm{total\ no.\ of \ questions}} \right)$. We report R@1, R@5, R@10 and average of the three is R-avg.

\subsection{Results \& Analysis}\label{subsec:result}
%
\begin{table*}[bp]
\centering
\resizebox{\textwidth}{!}{%
\begin{tabular}{|l|cccc|cccc|cccc|}
\hline
\multirow{2}{*}{Language} & \multicolumn{4}{c|}{\zeroshot}                                                                                               & \multicolumn{4}{c|}{\modelspeech}                                                                                            & \multicolumn{4}{c|}{\modeltext}                                                                                              \\ \cline{2-13} 
                          & \multicolumn{1}{c|}{R@1}           & \multicolumn{1}{c|}{R@5}           & \multicolumn{1}{c|}{R@10}          & R-avg         & \multicolumn{1}{c|}{R@1}           & \multicolumn{1}{c|}{R@5}           & \multicolumn{1}{c|}{R@10}          & R-avg         & \multicolumn{1}{c|}{R@1}           & \multicolumn{1}{c|}{R@5}           & \multicolumn{1}{c|}{R@10}          & R-avg         \\ \hline
Hindi (hi)                & \multicolumn{1}{c|}{19.8}          & \multicolumn{1}{c|}{38.5}          & \multicolumn{1}{c|}{51.6}          & 36.7          & \multicolumn{1}{c|}{\textbf{21.1}} & \multicolumn{1}{c|}{\textbf{46.5}} & \multicolumn{1}{c|}{\textbf{61.8}} & \textbf{43.1} & \multicolumn{1}{c|}{\textbf{40.9}} & \multicolumn{1}{c|}{\textbf{69.0}} & \multicolumn{1}{c|}{\textbf{79.4}} & \textbf{63.1} \\ \hline
Assamese (as)             & \multicolumn{1}{c|}{17.7}          & \multicolumn{1}{c|}{36.0}          & \multicolumn{1}{c|}{50.3}          & 34.7          & \multicolumn{1}{c|}{16.4}          & \multicolumn{1}{c|}{39.4}          & \multicolumn{1}{c|}{54.3}          & 36.7          & \multicolumn{1}{c|}{33.8}          & \multicolumn{1}{c|}{62.8}          & \multicolumn{1}{c|}{74.6}          & 57.1          \\ \hline
Bengali (bn)              & \multicolumn{1}{c|}{18.3}          & \multicolumn{1}{c|}{37.9}          & \multicolumn{1}{c|}{51.7}          & 36.0          & \multicolumn{1}{c|}{20.3}          & \multicolumn{1}{c|}{45.4}          & \multicolumn{1}{c|}{57.9}          & 41.2          & \multicolumn{1}{c|}{37.2}          & \multicolumn{1}{c|}{67.7}          & \multicolumn{1}{c|}{78.6}          & 61.2          \\ \hline
English (en)              & \multicolumn{1}{c|}{\textbf{21.5}} & \multicolumn{1}{c|}{\textbf{43.2}} & \multicolumn{1}{c|}{\textbf{55.9}} & \textbf{40.2} & \multicolumn{1}{c|}{16.9}          & \multicolumn{1}{c|}{40.4}          & \multicolumn{1}{c|}{53.7}          & 37.0          & \multicolumn{1}{c|}{37.2}          & \multicolumn{1}{c|}{64.5}          & \multicolumn{1}{c|}{78.0}          & 59.9          \\ \hline
Gujarati (gu)             & \multicolumn{1}{c|}{19.2}          & \multicolumn{1}{c|}{38.5}          & \multicolumn{1}{c|}{50.9}          & 36.2          & \multicolumn{1}{c|}{20.0}          & \multicolumn{1}{c|}{45.1}          & \multicolumn{1}{c|}{60.3}          & 41.8          & \multicolumn{1}{c|}{38.7}          & \multicolumn{1}{c|}{68.6}          & \multicolumn{1}{c|}{78.8}          & 62.0          \\ \hline
Kannada (kn)              & \multicolumn{1}{c|}{19.1}          & \multicolumn{1}{c|}{37.1}          & \multicolumn{1}{c|}{52.8}          & 36.3          & \multicolumn{1}{c|}{18.7}          & \multicolumn{1}{c|}{45.0}          & \multicolumn{1}{c|}{58.6}          & 40.8          & \multicolumn{1}{c|}{39.2}          & \multicolumn{1}{c|}{66.6}          & \multicolumn{1}{c|}{78.1}          & 61.3          \\ \hline
Malayalam (ml)            & \multicolumn{1}{c|}{18.5}          & \multicolumn{1}{c|}{35.1}          & \multicolumn{1}{c|}{49.1}          & 34.2          & \multicolumn{1}{c|}{18.3}          & \multicolumn{1}{c|}{44.7}          & \multicolumn{1}{c|}{57.8}          & 40.3          & \multicolumn{1}{c|}{35.0}          & \multicolumn{1}{c|}{65.2}          & \multicolumn{1}{c|}{77.4}          & 59.2          \\ \hline
Marathi (mr)              & \multicolumn{1}{c|}{19.0}          & \multicolumn{1}{c|}{36.2}          & \multicolumn{1}{c|}{52.3}          & 35.8          & \multicolumn{1}{c|}{18.7}          & \multicolumn{1}{c|}{44.8}          & \multicolumn{1}{c|}{59.6}          & 41.0          & \multicolumn{1}{c|}{38.4}          & \multicolumn{1}{c|}{65.7}          & \multicolumn{1}{c|}{77.0}          & 60.4          \\ \hline
Oriya (or)                & \multicolumn{1}{c|}{20.3}          & \multicolumn{1}{c|}{39.6}          & \multicolumn{1}{c|}{52.4}          & 37.4          & \multicolumn{1}{c|}{18.9}          & \multicolumn{1}{c|}{44.4}          & \multicolumn{1}{c|}{58.4}          & 40.5          & \multicolumn{1}{c|}{39.0}          & \multicolumn{1}{c|}{67.0}          & \multicolumn{1}{c|}{78.0}          & 61.4          \\ \hline
Punjabi (pa)              & \multicolumn{1}{c|}{19.5}          & \multicolumn{1}{c|}{37.9}          & \multicolumn{1}{c|}{50.5}          & 36.0          & \multicolumn{1}{c|}{20.3}          & \multicolumn{1}{c|}{45.7}          & \multicolumn{1}{c|}{58.5}          & 41.5          & \multicolumn{1}{c|}{39.2}          & \multicolumn{1}{c|}{67.4}          & \multicolumn{1}{c|}{78.3}          & 61.6          \\ \hline
Tamil (ta)                & \multicolumn{1}{c|}{18.9}          & \multicolumn{1}{c|}{38.5}          & \multicolumn{1}{c|}{52.0}          & 36.4          & \multicolumn{1}{c|}{17.7}          & \multicolumn{1}{c|}{43.8}          & \multicolumn{1}{c|}{59.0}          & 40.2          & \multicolumn{1}{c|}{34.9}          & \multicolumn{1}{c|}{66.5}          & \multicolumn{1}{c|}{77.5}          & 59.6          \\ \hline
Telugu (te)               & \multicolumn{1}{c|}{17.8}          & \multicolumn{1}{c|}{37.8}          & \multicolumn{1}{c|}{50.9}          & 35.5          & \multicolumn{1}{c|}{18.1}          & \multicolumn{1}{c|}{43.8}          & \multicolumn{1}{c|}{58.3}          & 40.1          & \multicolumn{1}{c|}{37.3}          & \multicolumn{1}{c|}{65.5}          & \multicolumn{1}{c|}{76.4}          & 59.7          \\ \hline
\end{tabular}%
}
\caption{Test set results using models trained with Hindi questions and evaluated on questions translated in 11 different languages.}
\label{tab:ab-lang}
\end{table*}

%

In Table~\ref{tab:main}, we present our main results. \model outperforms \zeroshot by an absolute 6.4\% indicating that our training method has successfully aligned speech features to the semantic space of LaBSE embeddings~\cite{labse}. \model is able to exploit very small amounts of weakly-annotated data to successfully bridge the modality gap and generate semantically-rich speech features. We also train \modeltext using ASR transcripts instead of speech using our architecture and we see large (absolute) 26.4\% and 19.98\% gains in R-avg performance compared to \zeroshot and \model, respectively. This suggests:
\begin{enumerate*}
    \item The quality of ASR transcripts is reasonably good with WER 17.8 for the base model without any LM (as reported by~\cite{javed2022towards}). 
    \item Aligning across speech and text modalities with small amounts of weakly labeled data is a challenging task. Bootstrapping using jointly trained speech and text models might be worth exploring as future work \cite{ao2021speecht5,chen2022maestro}.  
\end{enumerate*}
Our model \model still performs better than \zeroshot and can be quite effective when used with low-resource languages that do not have good ASR. We also perform experiments with other speech feature extractors such as Vakyansh~\cite{chadha2022vakyansh} and XLSR \cite{conneau2020unsupervised}, but both underperform compared to IndicWav2vec-Hindi~\cite{javed2022towards}.

\begin{table}[H]
\centering
\resizebox{\columnwidth}{!}{%
\begin{tabular}{|cc|c|c|c|c|}
\hline
\multicolumn{1}{|c|}{Feature extractor}         & Model Variant & R@1  & R@5  & R@10 & R-avg \\ \hline
\multicolumn{2}{|c|}{\zeroshot}                                 & 19.8 & 38.5 & 51.6 & 36.7  \\ \hline
\multicolumn{1}{|c|}{XLSR~\cite{conneau2020unsupervised}}                      & \model  & 9.3  & 30.5 & 43.8 & 27.9  \\ \hline
\multicolumn{1}{|c|}{\multirow{2}{*}{Vakyansh~\cite{chadha2022vakyansh}}} & \modeltext    & 35.6 & 62.9 & 74.9 & 57.8  \\ \cline{2-6} 
\multicolumn{1}{|c|}{}                          & \model  & 15.0 & 34.9 & 47.9 & 32.6  \\ \hline
\multicolumn{1}{|c|}{\multirow{2}{*}{IndicASR~\cite{javed2022towards}}} & \modeltext    & 40.9 & 69.0 & 79.4 & 63.1  \\ \cline{2-6} 
\multicolumn{1}{|c|}{}                          & \model  & 21.1 & 46.5 & 61.8 & 43.1  \\ \hline
\end{tabular}%
}
\caption{Main test results. We experiment with different feature extractors and model variants.} 
\label{tab:main}
\end{table}

A useful consequence of training \model using a frozen question encoder is that we can query our model with any of the 109 languages supported by LaBSE. We evaluate \model on 11 different Indian languages unseen during training. During inference, we translate the questionnaire using IndicTrans's Indic2English (for Hindi to English translation) and Indic2Indic (for Hindi to other Indic language translation) models~\cite{indictrans}. Note that during training we use speech in Hindi matched to questions in Hindi, while during inference we query using text in multiple Indian languages. From the results shown in Table~\ref{tab:ab-lang}, we observe that \model outperforms \zeroshot for all Indian languages. \modeltext performs the best, thus pointing to high-quality ASR transcriptions. Interestingly, performance on English is better with \zeroshot compared to \model. This could be because LaBSE was trained predominantly on English data. 

Tables ~\ref{tab:ab-N}, \ref{tab:ab-D}, \ref{tab:ab-B}, and \ref{tab:ab-sigma} show results on the dev set using \model from various ablation experiments by tuning important hyperparameters. Table~\ref{tab:ab-N} shows different group sizes. We see that having group sizes that are too large or too small are detrimental to performance. Note that each group contains an average of $14n$ chunks. We envisage that, with small groups there is not much diversity in negatives and with large groups a lot of potential negatives are never used since we only take minimum number of possible negatives across chunks in a batch (as described in Section~\ref{neg-sampling}). Hence, we train with $n=4$. 
\begin{table}[H]
\centering
\begin{tabular}{|c|c|c|c|c|}
\hline
$n$ & R@1           & R@5           & R@10          & R-avg         \\ \hline
2   & \textbf{18.7} & 43.4          & 59.5          & 40.5          \\ \hline
4   & 18.0          & \textbf{43.8} & \textbf{66.3} & \textbf{42.7} \\ \hline
8   & 18.4          & 44.5          & 59.9          & 40.9          \\ \hline
\end{tabular}
\caption{Tuning hyperparameter group size $n$, $|B| = 4, D=2, \sigma = 2.5$. Metrics reported on Dev set using \modelspeech.}
\label{tab:ab-N}
\end{table}
Data augmentation hyperparameter $D$ creates more groups of different segment combinations. Table ~\ref{tab:ab-D} shows how data augmentation helps, but at the cost of overfitting. $D=3$ leads to overfitting; $D=2$ gives the best trade-off between train and dev set performance.
\begin{table}[H]
\centering
\begin{tabular}{|c|c|c|c|c|}
\hline
$D$ & R@1           & R@5           & R@10          & R-avg         \\ \hline
1   & 13.4          & 39.3          & 57.9          & 36.9          \\ \hline
2   & \textbf{18.0} & 43.8          & \textbf{66.3} & 42.7          \\ \hline
3   & 16.9          & \textbf{46.1} & 65.3          & \textbf{42.8} \\ \hline
\end{tabular}
\caption{Tuning hyperparameter data augmentation $D$, $|B|=4, n=4, \sigma=2.5$. Metrics reported on Dev set using \modelspeech.}
\label{tab:ab-D}
\end{table}
In Table~\ref{tab:ab-B}, we present experiments with varying batch sizes. Similar to group size, we do not want the batches to be too large or too small. 
Although negative sampling is within a group, the number of negatives are the same within the batch and hence similar restrictions hold as with group sizes. Based on the results in Table~\ref{tab:ab-B}, we set batch size $|B|=4$. 
\begin{table}[H]
\centering
\begin{tabular}{|c|c|c|c|c|}
\hline
\textbf{$|B|$} & R@1           & R@5           & R@10          & R-avg         \\ \hline
2            & 16.1          & 43.0          & 60.4          & 39.8          \\ \hline
4            & \textbf{18.0} & \textbf{43.8} & \textbf{66.3} & \textbf{42.7} \\ \hline
8            & 17.0          & 42.6          & 61.6          & 40.4          \\ \hline
12           & 15.0          & 40.6          & 56.3          & 37.3          \\ \hline
\end{tabular}
\caption{Tuning hyperparameter batch size $|B|$, $n=4, D = 2, \sigma=2.5$. Metrics reported on the Dev set using \model.}
\label{tab:ab-B}
\end{table}
%
%
In Table~\ref{tab:ab-sigma}, we show experiments on tuning the standard deviation $\sigma$ of the Gaussian distribution from equation~\ref{eq:moving-G}. Using equation~\ref{eq:sigma}, we vary $\sigma$ and scale it with $\alpha$. In Table~\ref{tab:ab-sigma}, we present the metrics across various $\alpha$ values and constant $\sigma$ values. Our intuition behind using high $\sigma$ in the center was that central chunks are hard to anchor, but Table~\ref{tab:ab-sigma} suggests otherwise. We see that constant $\sigma$ is more effective by a significant margin, indicating that even central chunks are strongly anchored to some question. 
%
%
\begin{figure}[H]
    \centering
    \includegraphics[width=\columnwidth]{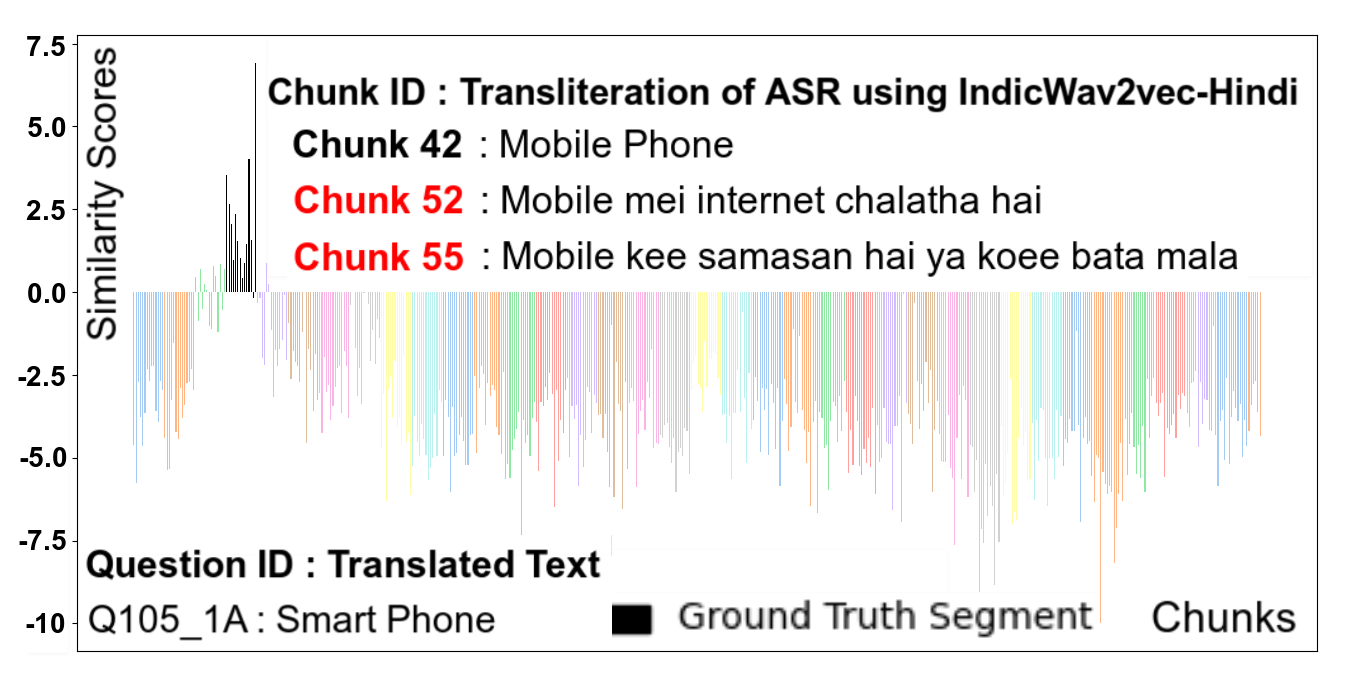}
    \caption{Chunk vs. question dot-product scores using \model. Same colored chunks belong to the same segment; black refers to ground-truth. Note that the question and the ASR transcript of the ground-truth segment are highly correlated. Chunks 52 and 55 have high non-negative scores due to the presence of the word \textit{mobile} and the question asked about \textit{smart phone} which matches exactly with chunk 42 \textit{mobile phone}.} 
    \label{fig:plot-good}
\end{figure}
\paragraph{Analysis.} We qualitatively analyze \model by plotting all the chunk scores for a given question. As shown in Figure~\ref{fig:plot-good}, \model is able to correctly identify the ground truth segment (in black); the dot-product scores are also very high for the ground truth segment compared to the rest.
\begin{table}[t!]
\centering
\begin{tabular}{|cc|c|c|c|c|}
\hline
\multicolumn{2}{|c|}{Standard deviation}                                                                  & R@1           & R@5           & R@10          & R-avg         \\ \hline
\multicolumn{1}{|c|}{\multirow{6}{*}{\begin{tabular}[c]{@{}c@{}}Fixed\\ $\sigma$\end{tabular}}} & 0.2  & 19.7          & \textbf{47.1} & 66.8          & 44.5          \\ \cline{2-6} 
\multicolumn{1}{|c|}{}                                                                             & 0.5  & \textbf{20.3} & 46.9          & \textbf{67.5} & \textbf{44.9} \\ \cline{2-6} 
\multicolumn{1}{|c|}{}                                                                             & 1    & 19.6          & 42.7          & 62.2          & 41.5          \\ \cline{2-6} 
\multicolumn{1}{|c|}{}                                                                             & 1.5  & 19.6          & 44.1          & 64.7          & 42.8          \\ \cline{2-6} 
\multicolumn{1}{|c|}{}                                                                             & 2.5  & 18.0          & 43.8          & 66.3          & 42.7          \\ \cline{2-6} 
\multicolumn{1}{|c|}{}                                                                             & 3    & 18.1          & 43.5          & 65.2          & 42.3          \\ \hline
\multicolumn{1}{|c|}{\multirow{6}{*}{\begin{tabular}[c]{@{}c@{}}Varying\\ $\alpha$\end{tabular}}}  & 0.1  & 15.8          & 41.8          & 57.3          & 38.3          \\ \cline{2-6} 
\multicolumn{1}{|c|}{}                                                                             & 0.25 & 15.9          & 42.4          & 58.8          & 39.0          \\ \cline{2-6} 
\multicolumn{1}{|c|}{}                                                                             & 0.4  & 18.0          & 43.7          & 60.0          & 40.6          \\ \cline{2-6} 
\multicolumn{1}{|c|}{}                                                                             & 0.6  & 17.9          & 43.6          & 63.0          & 41.5          \\ \cline{2-6} 
\multicolumn{1}{|c|}{}                                                                             & 0.75 & 18.1          & 43.0          & 65.2          & 42.1          \\ \cline{2-6} 
\multicolumn{1}{|c|}{}                                                                             & 0.9  & 17.5          & 40.4          & 63.6          & 40.5          \\ \hline
\end{tabular}
\caption{Metrics reported on Dev set with tuning $\sigma$ and $\alpha$; $|B|=4, n=4, D=2$ and \modelspeech architecture is used.}
\label{tab:ab-sigma}
\end{table}
\section{Conclusions and Future Work}
 We present a new end-to-end weakly-supervised approach \model, to align audio recordings with questions and an associated dataset of long health audio surveys collected from young mothers residing in rural areas of Bihar (India).
 \model automatically grounds questions within long audio recordings with the use of a Gaussian-weighted cross-attention mechanism that exploits the fact that questions appear temporally ordered in the training audio segments. We show through extensive experiments that we are able to isolate questions within long audio recordings reasonably well and also demonstrate how this framework can be used with queries in other Indian languages.

\appendix
\section*{Acknowledgments}
Unlinked anonymized data used for the work reported here was collected by CARE India Solutions for Sustainable Development, as part of system strengthening and governmental program implementation tracking and supported by Bill and Melinda Gates Foundation. The technical and legal support was provided by the Koita Centre  for Digital Health (KCDH), and specifically by Dr Raghavendran Lakshmi Narayanan, the Health Information Manager at KCDH. Shubham Nemani and Pranjal Saini's invaluable contributions and assistance during the initial data processing stages are deeply appreciated.

\bibliographystyle{named}
\bibliography{ijcai23}

\end{document}